\documentclass[letterpaper, 10 pt, conference]{ieeeconf}

\IEEEoverridecommandlockouts
\title{\LARGE \bf 6-DoF Object Pose from Semantic Keypoints
}

\author{Georgios Pavlakos$^{1}$,  Xiaowei Zhou$^{1}$,  Aaron Chan$^{1}$, Konstantinos G. Derpanis$^{2}$, and Kostas Daniilidis$^{1}$
\thanks{$^{1}$G. Pavlakos, X. Zhou, A. Chan and K. Daniilidis are with the  Department of Computer and Information Science, University of Pennsylvania, PA, USA, 
        {\tt\small \{pavlakos,xiaowz,aarchan,kostas\}@seas.upenn.edu}\newline
        $^{2}$K. Derpanis is with the Department of Computer Science, Ryerson University, ON, Canada,
        {\tt\small kosta@scs.ryerson.ca}}
}

\usepackage{algorithm}
\usepackage{algorithmic}
\usepackage{booktabs}
\usepackage{graphicx}
\usepackage{tabularx}
\usepackage{amsmath}
\usepackage{amssymb}
\usepackage{url}
\usepackage{subcaption}
\usepackage{color}
\usepackage{multirow}
\newcolumntype{P}[1]{>{\centering\arraybackslash}p{#1}}

\newcommand{\R}[1]{\mathbb{R}^{#1}}
\newcommand{\RR}[2]{\mathbb{R}^{#1 \times #2}}

\newcommand{\refEq}[1]{(\ref{#1})}

\def\bfc{{\boldsymbol{c}}}

\def\bfB{{\boldsymbol{B}}}

\def\bfD{{\boldsymbol{D}}}

\def\bfR{{\boldsymbol{R}}}
\def\bfS{{\boldsymbol{S}}}
\def\bfT{{\boldsymbol{T}}}

\def\bfW{{\boldsymbol{W}}}

\def\bfZ{{\boldsymbol{Z}}}

\def\bfzero{{\boldsymbol{0}}}
\def\bfone{{\boldsymbol{1}}}
\def\half{\frac{1}{2}~}

\begin{document}

\maketitle
\thispagestyle{empty}
\pagestyle{empty}

\begin{abstract}
This paper presents a novel approach to estimating the continuous six degree of freedom (6-DoF) pose (3D translation and rotation) of an object from a single RGB image. The  approach combines semantic keypoints predicted by a convolutional network (convnet) with a deformable shape model. Unlike prior work, we are agnostic to whether the object is textured or textureless, as the convnet learns the optimal representation from the available training image data. Furthermore, the approach can be applied to instance- and class-based pose recovery. Empirically, we show that the proposed approach can accurately recover the 6-DoF object pose for both instance- and class-based scenarios with a cluttered background. For class-based object pose estimation, state-of-the-art accuracy is shown on the large-scale PASCAL3D+ dataset.
\end{abstract}

\section{Introduction} 

\begin{figure}
  \centering
  \begin{subfigure}[b]{0.2\textwidth}
    \includegraphics[width=\linewidth]{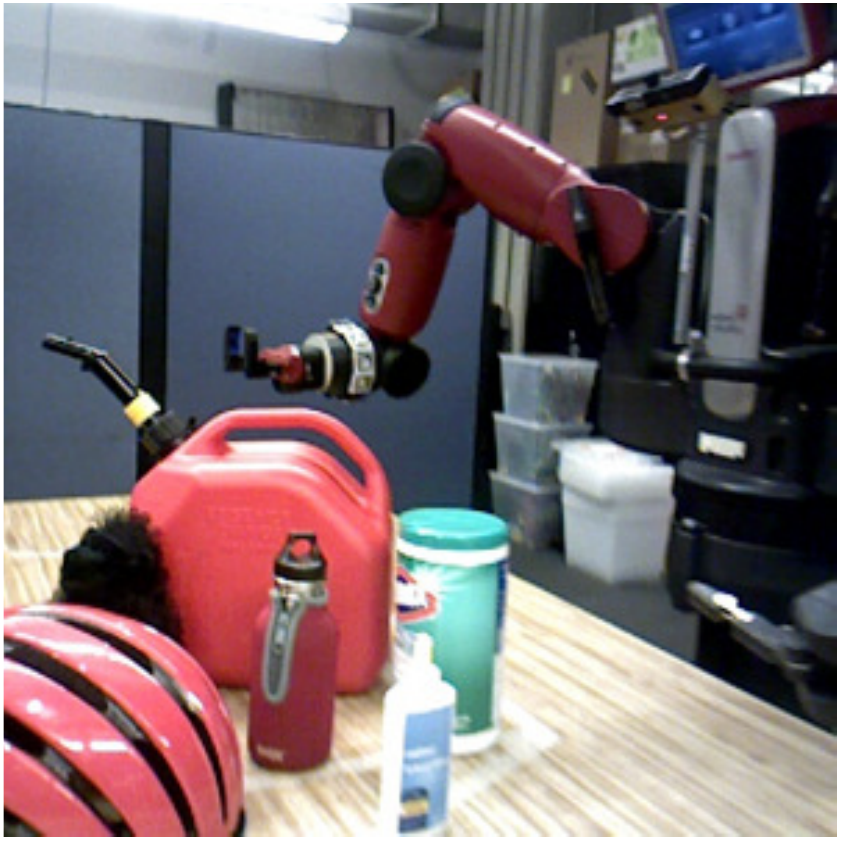}
  \end{subfigure}
  \begin{subfigure}[b]{0.2\textwidth}
  \includegraphics[width=\linewidth]{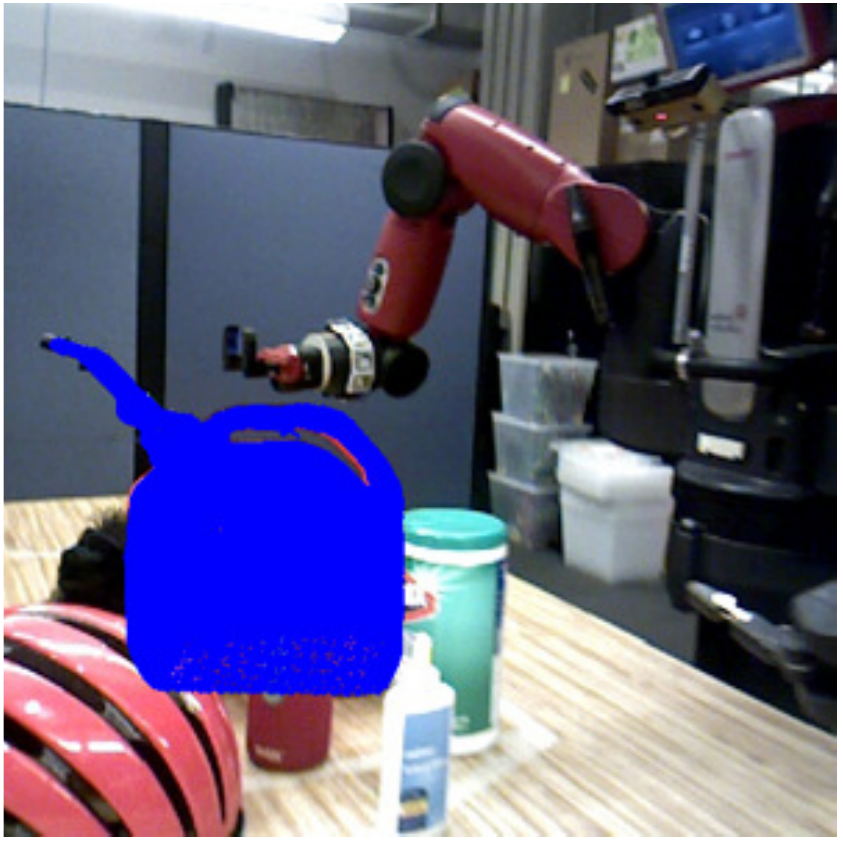}
  \end{subfigure}  
  \begin{subfigure}[b]{0.2\textwidth}
  \includegraphics[width=\linewidth]{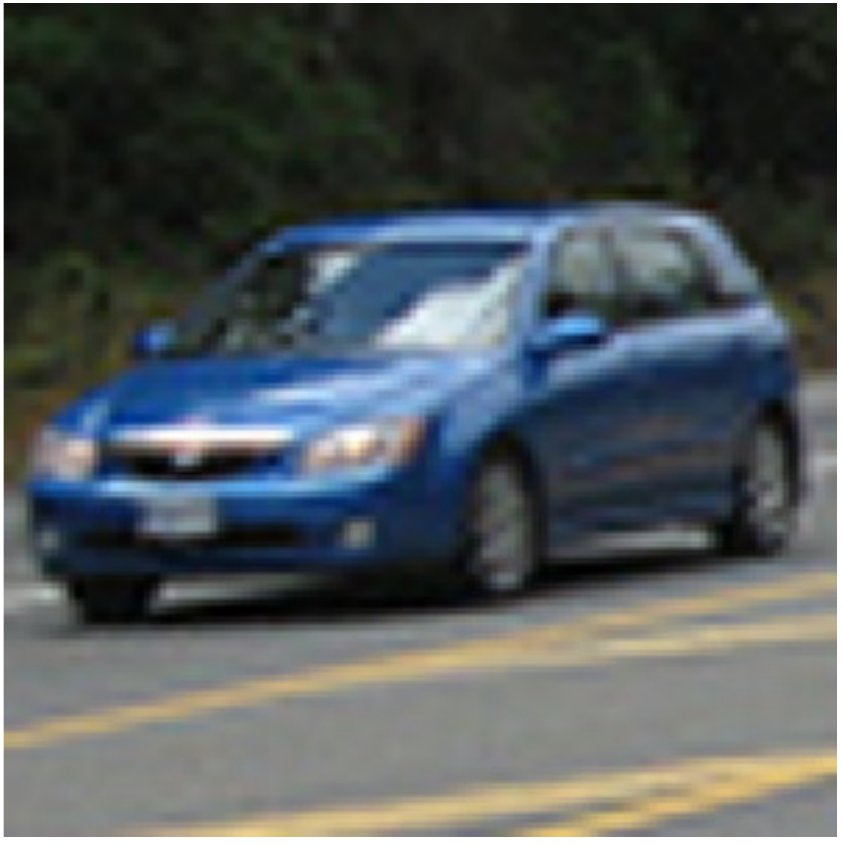}
  \end{subfigure}
  \begin{subfigure}[b]{0.2\textwidth}
  \includegraphics[width=\linewidth]{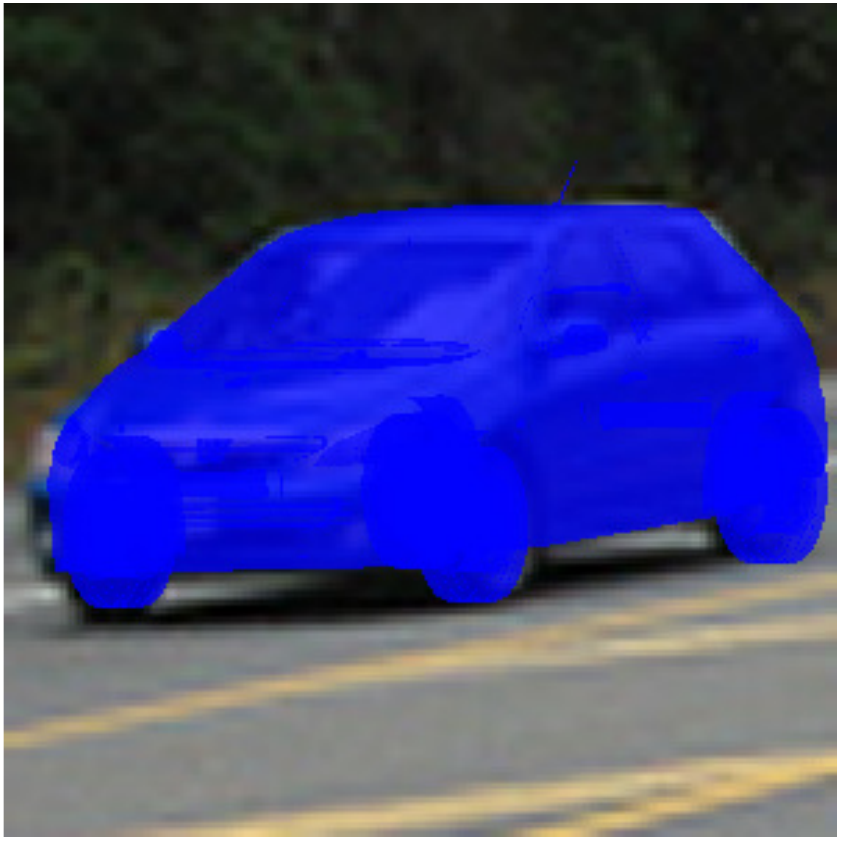}
 \end{subfigure}  
  \caption{Given a single RGB image of an object (left), we estimate its 6-DoF pose. 
The corresponding CAD model is overlaid on the image (right) using the estimated pose. 
Our method
deals with both instance-based (top) and class-based scenarios (bottom).
}\label{fig:intro}
\vspace{-2em}
\end{figure}

This paper addresses the task of estimating the continuous six degree of freedom (6-DoF) pose (3D translation and rotation) of an object from a single image.  Despite its importance in a variety of applications, e.g., robotic manipulation, and its intense study, most solutions tend to treat objects on a case-by-case basis.  For instance, approaches can be discerned by whether they address ``sufficiently'' textured objects with those that are textureless.  Some approaches focus on instance-based object detection while others address object classes. In this work, we strive for an approach where the admissibility of objects considered is as wide as possible (examples in Fig.~\ref{fig:intro}).
 
\begin{figure*}
  \centering
  \begin{subfigure}[b]{0.2\textwidth}
    \includegraphics[width=\linewidth]{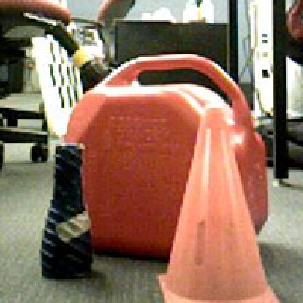}
    \caption{}
  \end{subfigure}
  \begin{subfigure}[b]{0.2\textwidth}
  \includegraphics[width=\linewidth]{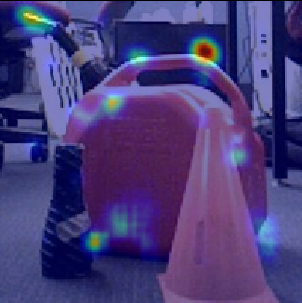}
  \caption{}
  \end{subfigure}  
  \begin{subfigure}[b]{0.2\textwidth}
  \includegraphics[width=\linewidth]{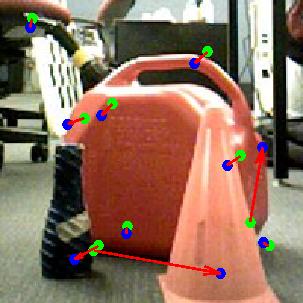}
  \caption{}
  \end{subfigure}
  \begin{subfigure}[b]{0.2\textwidth}
  \includegraphics[width=\linewidth]{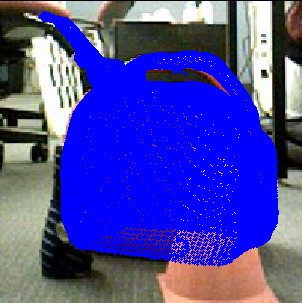}
  \caption{}
 \end{subfigure}  
  \caption{Pipeline of our approach. Given a single RGB image of an object (a), we localize a set of class-specific keypoints using a convnet with the stacked hourglass design. The output of this step is a set of heatmaps for each keypoint, which are combined for visualization in (b), sometimes leading to false detections. In (c), green dots represent the detected keypoints and the corresponding blue dots (connected with an arrow) the groundtruth locations. For robustness against such localization errors, we solve a fitting problem to enforce global consistency of the keypoints, where the response of the heatmaps is used as a measure of certainty for each keypoint. The optimization recovers the full 6-DoF pose of the object (d).}\label{fig:pipeline}
\vspace{-1em}
\end{figure*}

Our approach combines statistical models of appearance and the 3D shape layout of objects for pose estimation. It consists of two stages that first reasons about the 2D projected shape of an object captured by a set of 2D semantic keypoints and then estimates the 3D object pose consistent with the  keypoints. These steps are presented in Fig.~\ref{fig:pipeline}. In the first stage, we use a high capacity convolutional network (convnet) to predict a set of semantic keypoints. Here, the network takes advantage of its ability to aggregate appearance information over a wide-field of view, as compared to localized part models, e.g., \cite{gu2010}, to make reliable predictions of the semantic keypoints.  In the second stage, the semantic keypoint predictions are used to explicitly reason about the intra-class shape variability and the camera pose modeled by a weak or full perspective camera model. Pose estimates are realized by maximizing the geometric consistency between the parametrized deformable model and the 2D semantic keypoints. While this work focuses on RGB-based pose estimation, in the case where a corresponding point cloud is provided with the image, our method can provide a robust way to initialize the iterative closest point (ICP) algorithm \cite{ICP}, to further refine the pose.

\section{Related Work}
Estimating the 6-DoF pose of an object from a single image has attracted significant study. Given a rigid 3D object model and a set of 2D-to-3D point correspondences, various solutions have been explored, e.g., \cite{fischler1981, lepetit2009}. This is commonly referred to as the Perspective-n-Point problem (PnP). To relax the assumption of known 2D landmarks, a number of approaches \cite{collet2009,collet2011,xie2013} have considered the detection of discriminative image keypoints, such as SIFT \cite{lowe2004}, with highly textured objects.  A drawback with these approaches is that they are inadequate for addressing textureless objects and  their performance is susceptible to scene clutter. An alternative to sparse discriminative keypoints is offered by dense methods~\cite{brachmann2014learning,brachmann2016uncertainty,doumanoglou2016recovering,michel2016global}, where every pixel or patch is voting for the object pose. These approaches are also applicable for textureless objects, however, the assumption that a corresponding instance-specific 3D model is available for each object limits their general applicability.

Holistic template-based approaches are one of the earliest approaches considered in the object detection literature. To accommodate appearance variation due to camera capture viewpoint, a set of template images of the object instance are captured about the view sphere and are compared to the input image at runtime. In recent years, template-based methods have received renewed interest due to the advent of accelerated matching schemes and their ability to detect textureless objects by way of focusing their model description on the object shape \cite{muja2011,Hinterstoisser2012,rios2013,xie2013,cao2016}. While impressive results in terms of accuracy and speed have been demonstrated, holistic template-based approaches are limited to instance-based object detection. To address class variability and viewpoint, various approaches have used a collection of 2D appearance-based part templates trained separately on discretized views \cite{gu2010,fidler2012,pepik2012,xiang2014,zhu2014single}.  

Convolutional networks (convnets) \cite{lecun1989,krizhevsky2012} have emerged as the method of choice for a variety of problems.  Closest to the current work is their application in camera viewpoint and keypoint prediction. Convnets have been used to predict the camera's viewpoint with respect to the object by way of direct regression or casting the problem as classification into a set discrete views \cite{massa2014,tulsiani2015vk,su2015}.  While these approaches allow for object category pose estimation they do not provide fine-grained information about the 3D layout of the object. Convnet-based keypoint prediction for human pose estimation (e.g., \cite{toshev2014,zhou2015sparseness,newell2016stacked,wei2016cpm}) has attracted considerable study, while limited attention has been given to their application with generic object categories \cite{long2014,tulsiani2015vk}. Their success is due in part to the high discriminative capacity of the network.  Furthermore, their ability to aggregate information over a wide field of view allows for the resolution of ambiguities (e.g., symmetry) and for localizing occluding joints.

Statistical shape-based models tackle recognition by aligning a shape subspace model to image features.  While originally proposed in the context of 2D shape  \cite{cootes1995} they have proven useful for modelling the 3D shape of a host of object classes, e.g., faces \cite{cao20133d}, cars \cite{zia2013detailed,murthy2016reconstructing} and human pose \cite{ramakrishna2012}.  In recent work \cite{zhu_popup}, data-driven discriminative landmark hypotheses were combined with a 3D deformable shape model and a weak perspective camera model in a convex optimization framework to globally recover the shape and pose of an object in a single image.  Here, we adapt this approach and extend it with a perspective camera model, in cases where the camera intrinsics are known.

\vspace{5pt}\noindent{\bf Contributions }In the light of previous work, the contributions of our work are as follows:
\begin{itemize}
\item We present an efficient approach that combines highly reliable (semantic) keypoints predicted by a convnet with a deformable shape model to estimate the continuous 6-DoF pose of an object. Unlike previous work, we are agnostic to whether the object is textured or textureless, as the convnet learns the optimal representation from the available image training data.  Furthermore, the same approach can be applied to instance- and class-based pose recovery.  

\item Empirically, we demonstrate that the proposed approach yields accurate 6-DoF pose estimates in scenes with cluttered backgrounds without the requirement of any pose initialization. State-of-the-art performance is shown on the large-scale PASCAL3D+ dataset \cite{xiang2014}.  
\end{itemize}

\section{Technical Approach}
The proposed pipeline includes object detection, keypoint localization and pose optimization. As object detection has been a well studied problem, we assume that a bounding box around the object has been provided by an off-the-shelf object detector, e.g., Faster R-CNN \cite{ren2015faster}, and focus on the keypoint localization and pose optimization.

\subsection{Keypoint localization}
The keypoint localization step employs the ``stacked hourglass'' network architecture~\cite{newell2016stacked} that has been shown to be particularly effective for 2D human pose estimation. Motivated by this success, we use the same network design and train the network for object keypoint localization. 

\begin{figure*}
  \centering
  \includegraphics[width=0.8\linewidth,trim={5cm 5cm 5cm 5cm},clip]{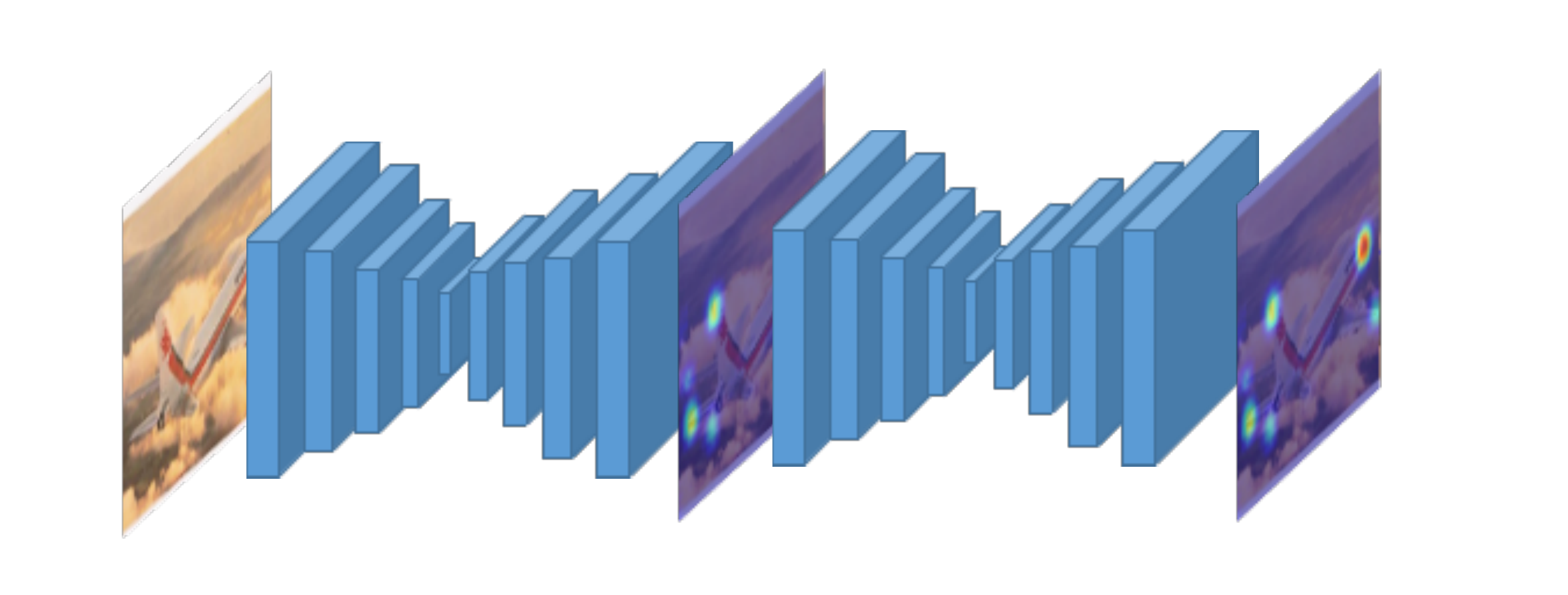}\\
  \vspace{-2em}
	  \begin{tabular}{P{4.5cm} P{4.5cm} P{5cm}} 
	  \scriptsize{Image} & \scriptsize{Intermediate heatmaps} & \scriptsize{Output heatmaps} \\ 
	  \end{tabular}
  \caption{Overview of the stacked hourglass architecture. Here, two hourglass modules are stacked together. The symmetric nature of the design allows for bottom-up processing (from high to low resolution) in the first half of the module, and top-down processing (from low to high resolution) in the second half. Intermediate supervision is applied after the first module. The heatmap responses of the second module represent the final output of the network that is used for keypoint localization.}\label{fig:hourglass}
\vspace{-1em}
\end{figure*}

\vspace{5pt}\noindent{\bf Network architecture }
A high level overview of the main network components is presented in Fig. \ref{fig:hourglass}. The network takes as input an RGB image, and outputs a set of heatmaps, one per keypoint, with the intensity of the heatmap indicating the confidence of the respective keypoint to be located at this position. The network consists of two hourglass components, where each component can be further subdivided into two main processing stages. In the first stage, a series of convolutional and max-pooling layers are applied to the input. After each max-pooling layer, the resolution of the feature maps decreases by a factor of two, allowing the next convolutional layer to process the features at a coarser scale. This sequence of processing continues until reaching the lowest resolution ($4\times4$ feature maps), which is illustrated by the smallest layer in the middle of each module in Fig. \ref{fig:hourglass}. Following these downsampling layers, the processing continues with a series of convolutional and upsampling layers. Each upsampling layer increases the resolution by a factor of two. This process culminates with a set of heatmaps at the same resolution as the input of the hourglass module. A second hourglass component is stacked at the end of the first one to refine the output heatmaps. The groundtruth labels used to supervise the training are synthesized heatmaps based on a 2D Gaussian centered at each keypoint with a standard deviation set to one. The $\ell_2$ loss is minimized during training. Optionally, intermediate supervision can be applied at the end of the first module, which provides a richer gradient signal to the network and guides the learning procedure towards a better optimum \cite{lee2015deeply}. The heatmap responses of the last module are considered as the final output of the network and the peak in each heatmap indicates the most likely location for the corresponding keypoint. 

\vspace{5pt}\noindent{\bf Design benefits }
The most critical design element of the hourglass network is the symmetric combination of bottom-up and top-down processing that each hourglass module performs. Given the large appearance changes of objects due to in-class and viewpoint variation, both local and global cues are needed to effectively decide the locations of the keypoints in the image. The consolidation of features across different scales in the hourglass architecture allows the network to successfully integrate both local and global appearance information, and commit to a keypoint location only after this information has been made available to the network. Moreover, the stacking of hourglass modules provides a form of iterative processing that has been shown to be effective with several other recent network designs~\cite{carreira2015iterative,wei2016cpm} and offers additional refinement of the network estimates. Additionally, the application of intermediate supervision at the end of each module has been validated as an effective training strategy, particularly ameliorating the practical issue of vanishing gradients when training a deep neural network \cite{lee2015deeply}. Finally, residual layers are introduced~\cite{he2015residual}, which have achieved state-of-the-art results for many visual tasks, including object classification~\cite{he2015residual}, instance segmentation~\cite{dai2015instance}, and 2D human pose estimation ~\cite{newell2016stacked}.

\subsection{Pose optimization}\label{sec:optimization}
Given the keypoint locations on the 3D model as well as their correspondences in the 2D image, one naive approach is to simply apply an existing PnP algorithm to solve for the 6-DoF pose. This approach is problematic because the keypoint predictions by the convnet can be rendered imprecise due to occlusions and false detections in the background. Moreover, the exact 3D model of the object instance in the testing image is often unavailable. To address these difficulties, we propose to fit a deformable shape model to the 2D detections
 while considering the uncertainty in keypoint predictions.   

A deformable shape model is built for each object category using 3D CAD models with annotated keypoints. More specifically, the $p$ keypoint locations on a 3D object model are denoted by $\bfS\in\RR{3}{p}$ and
\begin{align}\label{eq:shape-model}
    \bfS = \bfB_0 + \sum_{i=1}^{k} c_i\bfB_i,
\end{align}
where $\bfB_0$ is the mean shape of the given 3D model and $\bfB_1,\dotsc,\bfB_k$ are several modes of possible shape variability computed by Principal Component Analysis (PCA).

Given detected keypoints in an image, which are denoted by $\bfW\in\RR{2}{p}$, the goal is to estimate the rotation $\bfR\in\RR{3}{3}$ and translation $\bfT\in\RR{3}{1}$ between the object  and camera frames as well as the coefficients of the shape deformation $\bfc=[c_1,\cdots,c_k]^\top$. 

The inference is formulated as the following optimization problem:
\begin{align}\label{eq:cost}
    \min_{\theta} ~~ & \half \left\| \xi(\theta)\bfD^{\frac{1}{2}} \right\|_F^2 + \frac{\lambda}{2} \|\bfc\|_2^2,
\end{align}
where $\theta$ is the set of unknowns, $\xi(\theta)$ denotes the fitting residuals dependent on $\theta$, 
and the Tikhonov regularizer $\|\bfc\|_2^2$ is introduced to penalize large deviations from the mean shape. 

To incorporate the uncertainty in 2D keypoint predictions, a diagonal weighting matrix $\bfD\in\RR{p}{p}$ is introduced:
\begin{align}
\bfD &= \begin{bmatrix}
            d_1 & 0 & \cdots & 0\\
            0 & d_2 & \cdots & 0\\
            \vdots & \vdots & \ddots & \vdots\\
            0 & 0 & \cdots & d_p\\
            \end{bmatrix},
\end{align}
where $d_{i}$ indicates the localization confidence of the $i$th keypoint in the image. In our implementation, $d_{i}$ is assigned the peak value in the heatmap corresponding to the $i$th keypoint. As shown previously \cite{newell2016stacked}, the peak intensity of the heatmap provides a good indicator for the visibility of a keypoint in the image.

The fitting residuals, $\xi(\theta)$, measure the differences between the given 2D keypoints, provided by the previous processing stage, and the projections of 3D keypoints. Two camera models are next considered.

\subsubsection{Weak perspective model}

If the camera intrinsic parameters are unknown, the weak perspective camera model is adopted, which is usually a good approximation to the full perspective case when the camera is relatively far away from the object. In this case, the reprojection error is written as 
\begin{align}
    \xi(\theta) = \bfW - s \bar{\bfR}\left(\bfB_0 + \sum_{i=1}^{k} c_i\bfB_i\right) - \bar{\bfT}\bfone^\top,
\end{align}
where $s$ is a scalar, $\bar{\bfR}\in\RR{2}{3}$ and $\bar{\bfT}\in\R{2}$ denote the first two rows of $\bfR$ and $\bfT$, respectively, and $\theta=\{s,\bfc,\bar{\bfR},\bar{\bfT}\}$.

The problem in \refEq{eq:cost} is continuous and in principal can be locally solved by any gradient-based method. We solve it with a block coordinate descent scheme because of its fast convergence and the simplicity in implementation. We alternately update each of the variables while fixing the others. The updates of $s$, $\bfc$ and $\bar{\bfT}$ are simply solved using closed-form least squares solutions. The update of $\bar{\bfR}$ should consider the $SO(3)$ constraint. Here, the Manopt toolbox \cite{boumal2014manopt} is used to optimize $\bar{\bfR}$ over the Stiefel manifold. As the problem in \refEq{eq:cost} is non-convex, we further adopt a convex relaxation approach \cite{zhou20153d} to initialize the optimization. More specifically, we only estimate the pose parameters while fixing the 3D model as the mean shape in the initialization stage. By setting $\bfc=\bfzero$ and replacing the orthogonality constraint on $\bar{\bfR}$ by the spectral norm regularizer, the problem in \refEq{eq:cost} can be converted to a convex program and solved with global optimality \cite{zhou20153d}.

\subsubsection{Full perspective model}

If the camera intrinsic parameters are known, the full perspective camera model is used, and the residuals are defined as
\begin{align}
\xi(\theta) = \tilde{\bfW}\bfZ- {\bfR}\left(\bfB_0 + \sum_{i=1}^{k} c_i\bfB_i\right) - {\bfT}\bfone^\top,
\end{align}
where $\tilde{\bfW}\in\RR{3}{p}$ represents the normalized homogeneous coordinates of the 2D keypoints and $\bfZ$ is a diagonal matrix: 
\begin{align}
\bfZ &= \begin{bmatrix}
            z_1 & 0 & \cdots & 0\\
            0 & z_2 & \cdots & 0\\
            \vdots & \vdots & \ddots & \vdots\\
            0 & 0 & \cdots & z_p\\
            \end{bmatrix},
\end{align}
where $z_i$ is the depth for the $i$th keypoint in 3D. Intuitively, the distances from the 3D points to the rays crossing the corresponding 2D points are minimized. In this case, the unknown parameter set $\theta$ is given by $\{\bfZ,\bfc,\bfR,\bfT\}$.

The optimization here is similar to the alternating scheme in the weak perspective case. The update of $\bfZ$ also admits a closed-form solution and the update of $\bfR$ can be analytically solved by the orthogonal Procrustes analysis. To avoid local minima, the optimization is initialized by the weak perspective solution.  

\section{Experiments}
\subsection{Instance-based pose recovery: gas canister}

This section considers the recovery of pose for a specific object instance. This case fits well with many robotics applications where the objects in the environment are known. Moreover, it allows us to establish the accuracy of our approach in a relatively simple setting before dealing with the more challenging object class scenario.

We collected a dataset of 175 RGB-D images of a textureless gas canister. 
The depth data was only used to generate the groundtruth. More specifically, a complete 3D model of the gas canister was reconstructed using KinectFusion \cite{newcombe2011} and the groundtruth object pose for each image was calculated by ICP with careful manual initialization. Then, 10 keypoints were manually defined on the 3D model and projected to the images yielding groundtruth keypoint locations in 2D for training the convnet. 

\begin{figure*}{}
  \centering
  \includegraphics[width=0.8\linewidth]{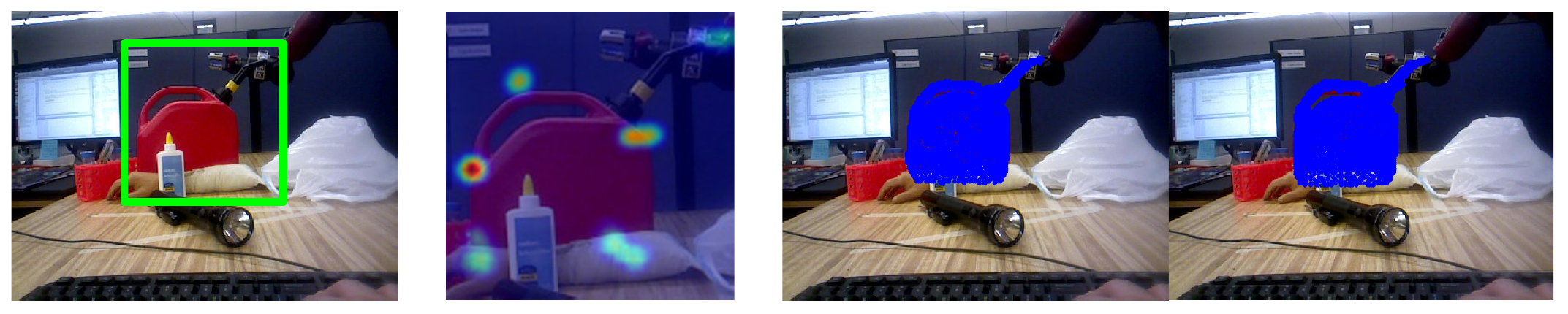}\\
  \includegraphics[width=0.8\linewidth]{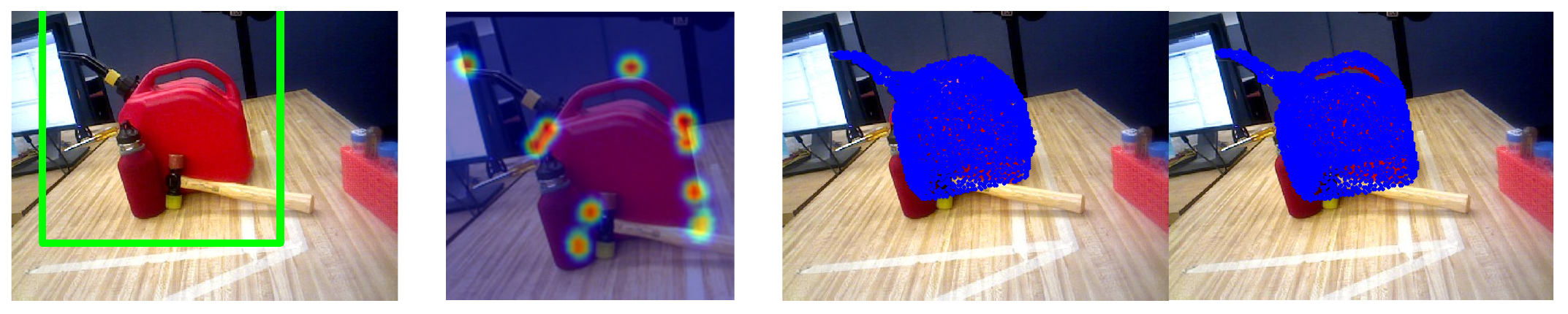}\\
  \includegraphics[width=0.8\linewidth]{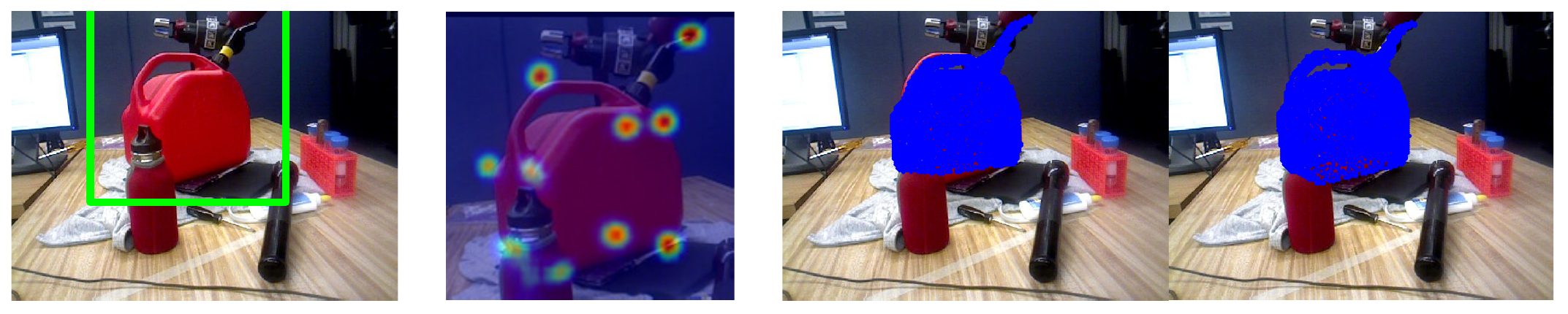}\\
  \caption{Qualitative results on the gas canister dataset. From left-to-right: RGB images with bounding boxes provided by Faster R-CNN~\cite{ren2015faster}, heatmaps from the convnet, projections of the 3D model with estimated poses using the weak-perspective model and full-perspective model, respectively. Note the better alignment near the handle with the full-perspective model.}\label{fig:gascan}
\end{figure*}

A random 85\%/15\% split was used for the training/test data. A stacked hourglass network with two hourglass modules was trained. The output heatmaps for the testing images are visualized in the second column of Fig. \ref{fig:gascan}.  As can be seen, the hourglass network is able to locate the keypoints reliably in the presence of viewpoint variety and occlusions. The non-visible keypoints are also well localized due to the network's ability to take global context into account. The estimated object poses are shown in the last two columns of Fig. \ref{fig:gascan}. The projected 3D models align accurately with the image; the full-perspective solution is more precise than the weak-perspective one. It is worth noting that only 150 images were used to train the network from scratch. Overfitting might be an issue with such a small training set, but the empirical results suggest that the hourglass model captures the object appearance very well in this single instance case. More challenging examples with large intra-class variability are considered in Section \ref{sec:pascal}. 

The 6-DoF pose was estimated with the known 3D model and camera intrinsic parameters using the optimization in Section \ref{sec:optimization}. The following geodesic distance was used to measure the rotation error between a pose estimate, $R_1$, and the groundtruth, $R_2$: 
\begin{align}
\Delta(R_1,R_2) = \frac{\|\log(R_1^TR_2)\|_{F}}{\sqrt{2}}.\label{eq:geodesic}
\end{align}
As a simple baseline, the following greedy approach was implemented: the maximum response locations in the heatmaps were selected as 2D keypoint locations and the standard PnP problem was solved by the EPnP algorithm~\cite{lepetit2009} to estimate the object pose. The results are presented in Table~\ref{tab:gas}. While the weak-perspective solutions (Proposed WP) are on average worse than EPnP due to the inaccurate camera model, the full-perspective solutions (Proposed FP) are much more precise than those of EPnP.  The remarkably small pose errors returned by the proposed approach based on a single RGB image are in the range suitable for a general grasping system. In the supplemental video\footnote{https://www.seas.upenn.edu/\%7Epavlakos/projects/object3d}, we demonstrate the efficacy of our approach in this situation by way of a robot grasping experiment. 

\begin{table}
\caption{Pose estimation errors on the gas canister dataset.}
\centering
\renewcommand{\arraystretch}{1.2}
\begin{tabular}{c*{15}{c}}
\hline
 \multirow{2}{*}{Approach}& \multicolumn{2}{c}{Rotation (degree)} & \multicolumn{2}{c}{Translation (mm)} \\
& Mean & Median & Mean & Median \\
\hline
Proposed WP & 7.99 & 7.61 & N/A & N{}/A \\
Proposed FP & 3.57 & 3.11 & 12.05 & 8.82 \\
EPnP \cite{lepetit2009} & 7.17 & 5.21 & 43.45 & 21.51 \\
\hline
\end{tabular}
\vspace{0.25em}
\label{tab:gas}
\end{table}{}

\subsection{Class-based pose recovery: PASCAL3D+}\label{sec:pascal}

Moving to a more challenging scenario, we demonstrate the full strength of our approach using the large-scale PASCAL3D+ dataset \cite{xiang2014}. The stacked hourglass network was trained from scratch with the training set of PASCAL3D+. Instead of training separate models for different object classes, a single network was trained to output heatmap predictions for all of the 124 keypoints from all classes. Using a single network for all keypoints allows us to share features across the available classes and significantly decreases the number of parameters needed for the network. At test time, given the class of the test object, the heatmaps corresponding to the keypoints belonging to this class were extracted. For pose optimization, two cases were tested: (i) the CAD model for the test image was known; and (ii) the CAD model was unknown and the pose was estimated with a deformable model whose basis was learned by PCA on all CAD models for each class in the dataset. Two principal components were used ($k=2$) for each class, which was sufficient to explain greater than $95\%$ of the shape variation. The 3D model was fit to the 2D keypoints with a weak-perspective model as the camera intrinsic parameters were not available. 

\begin{figure*}
  \centering
  \includegraphics[width=\linewidth]{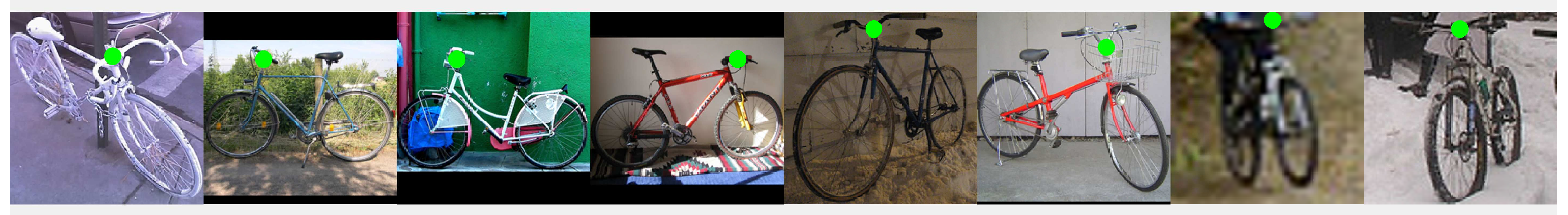}\\
  \includegraphics[width=\linewidth]{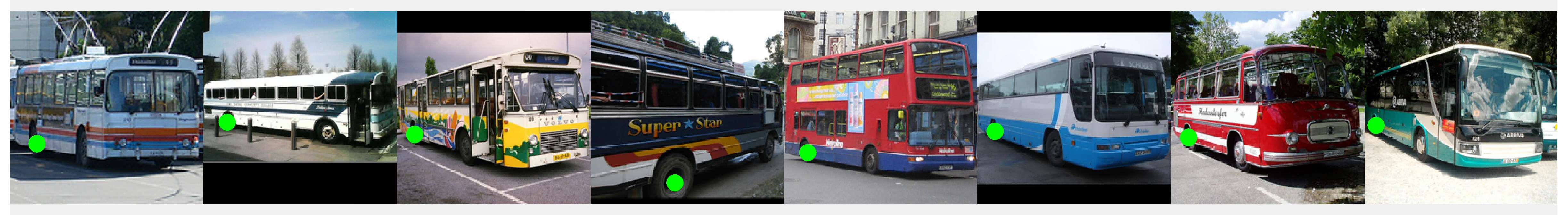}\\
  \includegraphics[width=\linewidth]{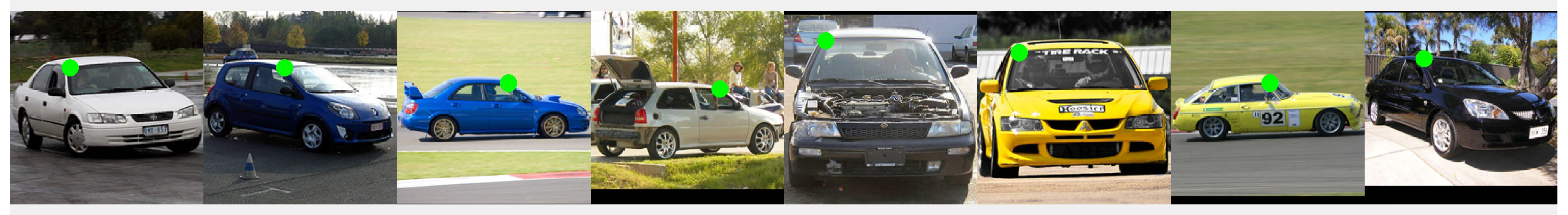}\\
  \includegraphics[width=\linewidth]{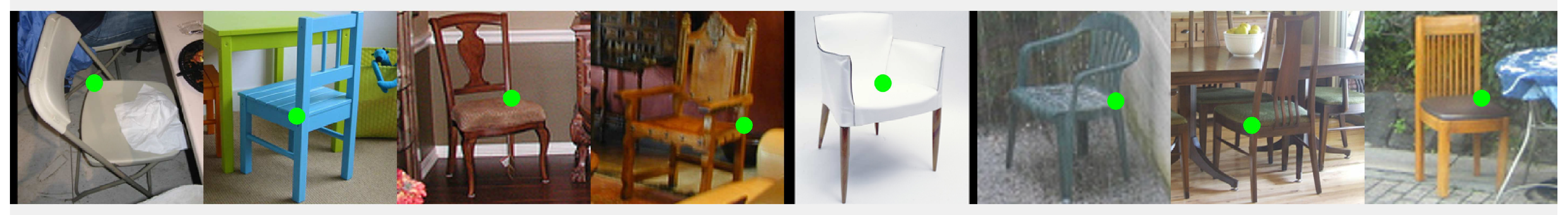}\\
  \caption{Localization results for diverse keypoint categories. We visualize eight images selected randomly from the top 50 responses for each keypoint. The keypoint localization network is particularly successful at establishing semantic correspondences across the instances of a class, despite the significant intra-class variation and wide ranging camera viewpoints.}\label{fig:semantic}
\end{figure*}

\vspace{5pt}\noindent{\bf Semantic correspondences } A crucial component of our approach is the powerful learning procedure that is particularly successful at establishing correspondences across the semantically related keypoints of each class. To demonstrate this network property,  in Fig.~\ref{fig:semantic} we present a subset of the keypoints for each class along with the localizations of these keypoints in a randomly selected set of images among the ones with the top 50 responses. It is interesting to note that despite the large appearance differences due to extreme viewpoint and intra-class variability, the predictions are very consistent and preserve the semantic relation across various class instances.

\begin{figure*}
  \centering
  \includegraphics[width=0.49\linewidth]{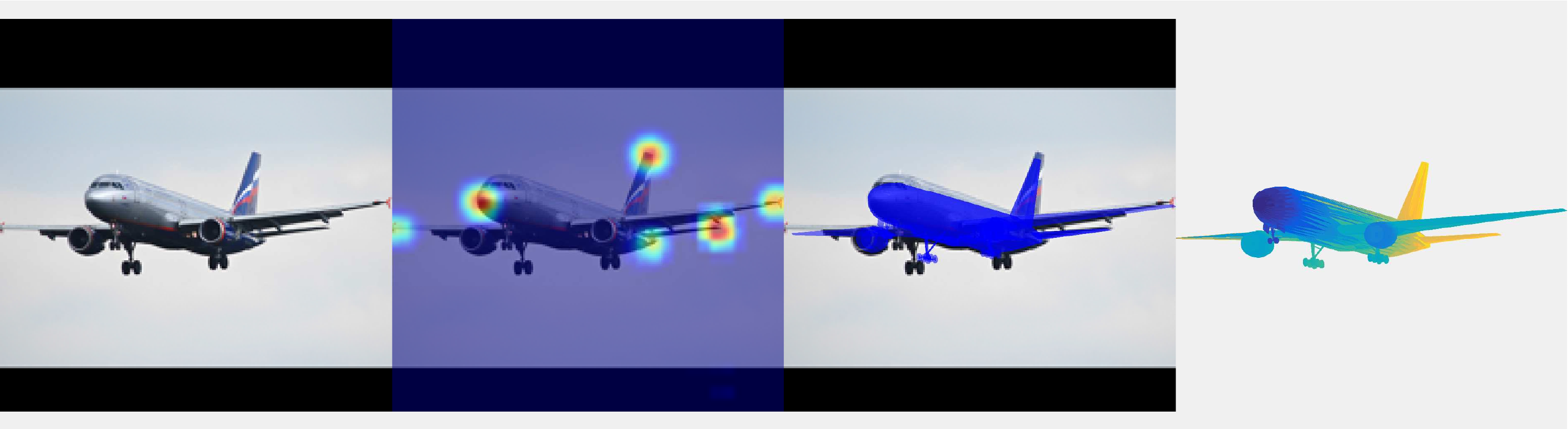}
  \includegraphics[width=0.49\linewidth]{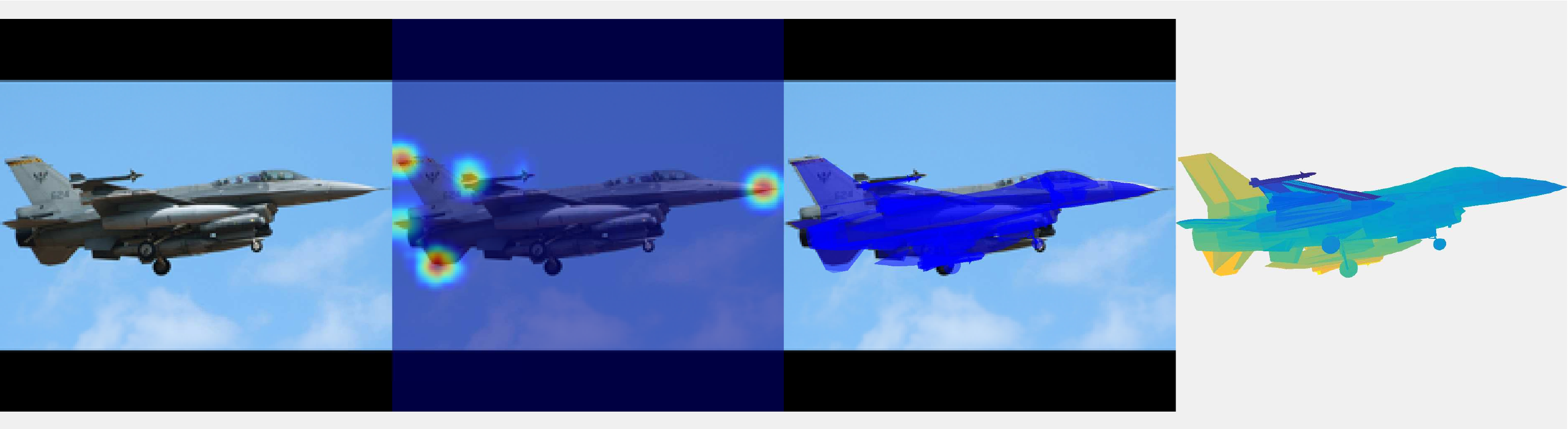}
  \includegraphics[width=0.49\linewidth]{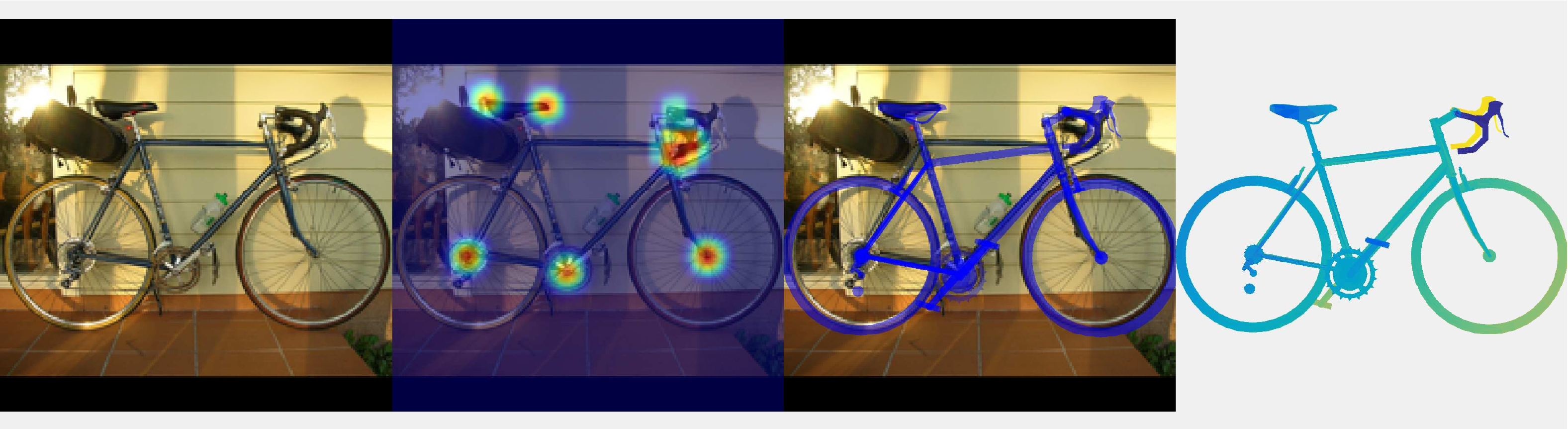}
  \includegraphics[width=0.49\linewidth]{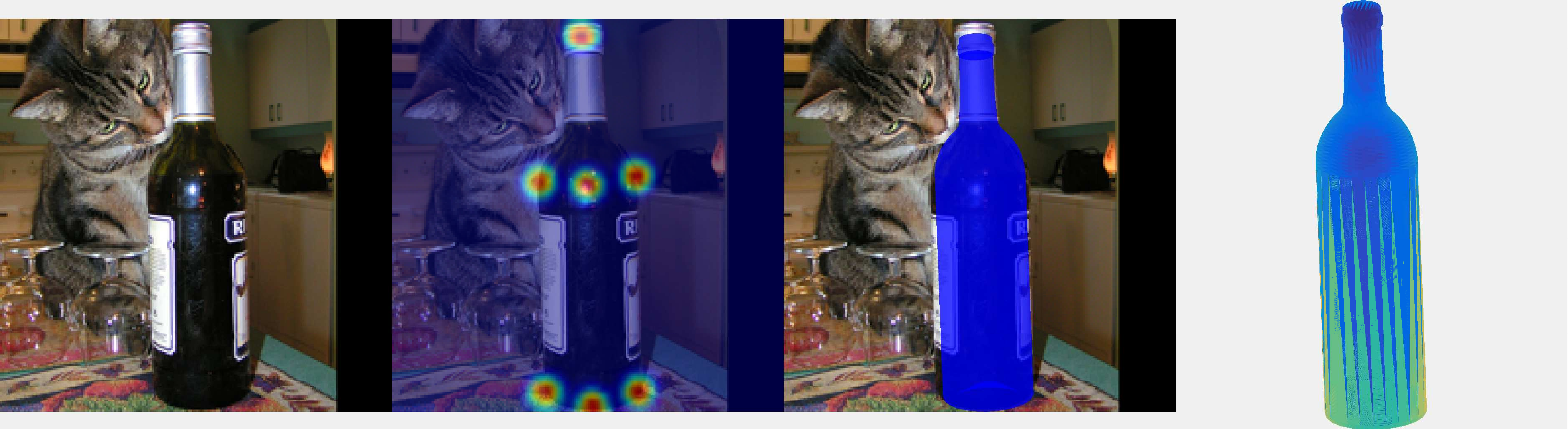}
  \includegraphics[width=0.49\linewidth]{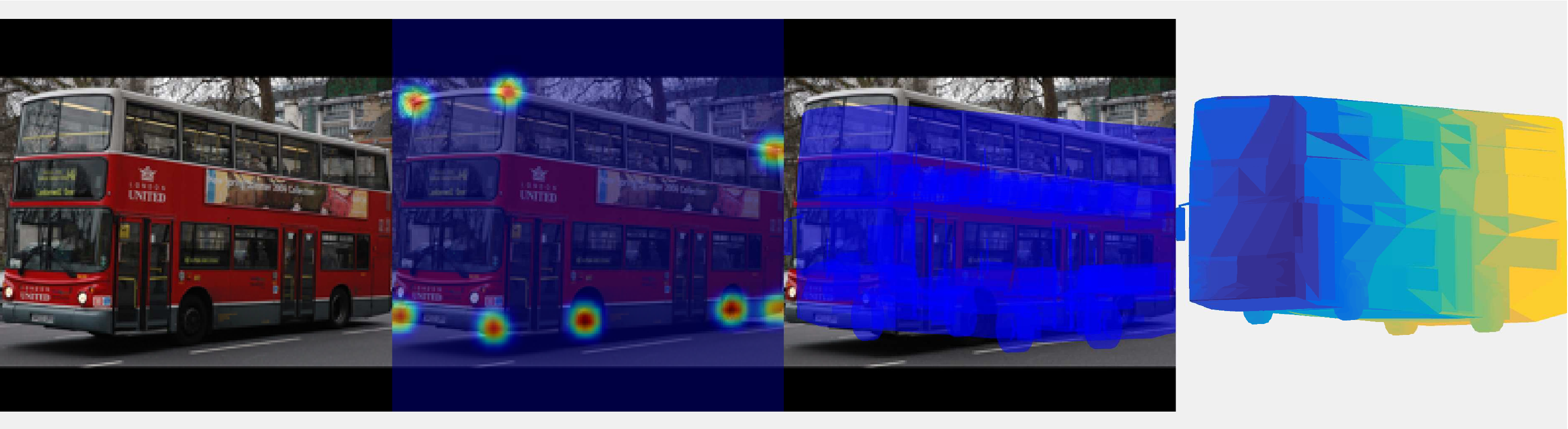}
  \includegraphics[width=0.49\linewidth]{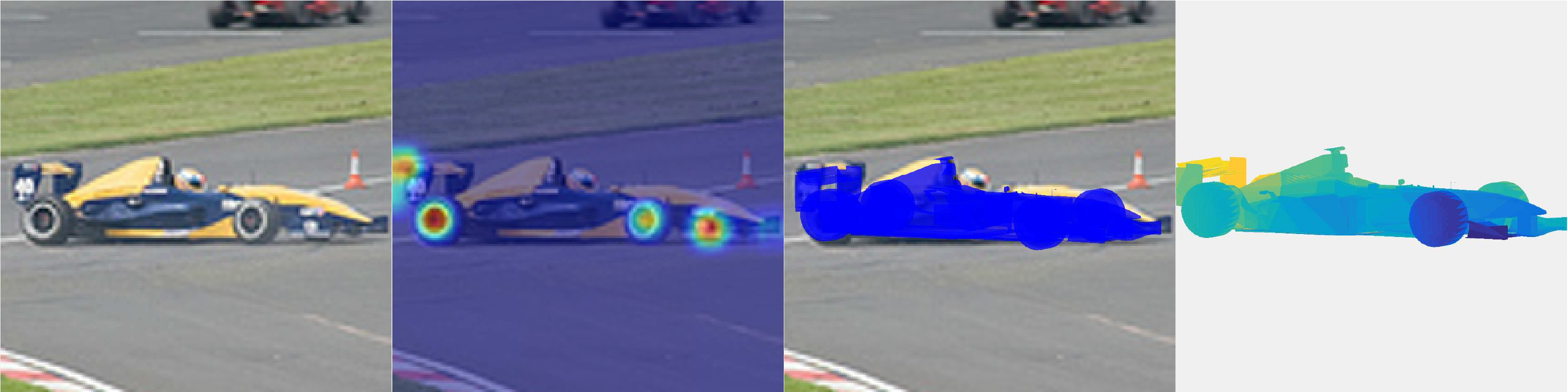}
  \includegraphics[width=0.49\linewidth]{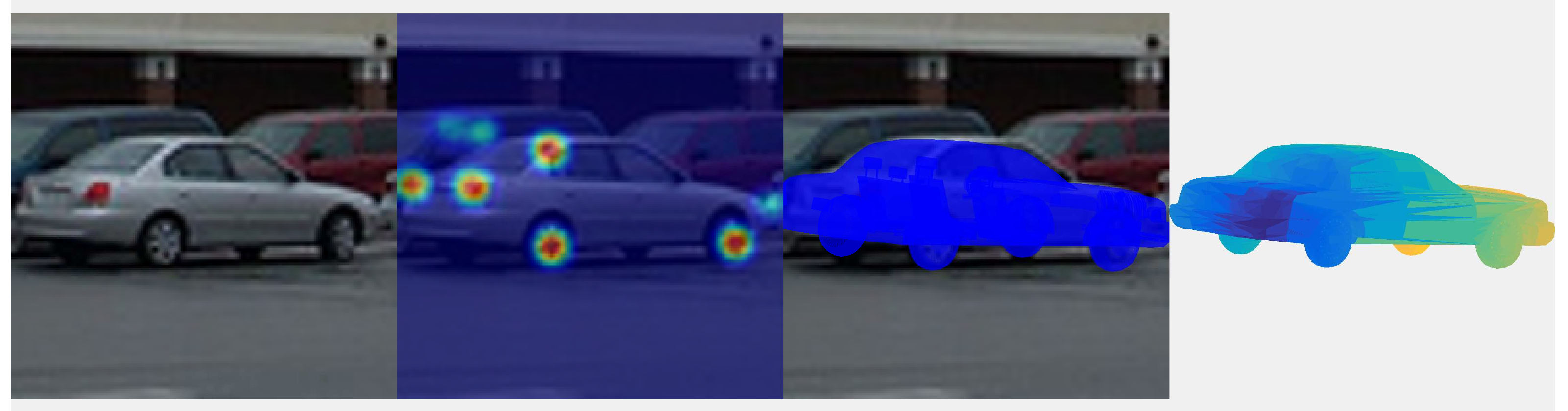}
  \includegraphics[width=0.49\linewidth]{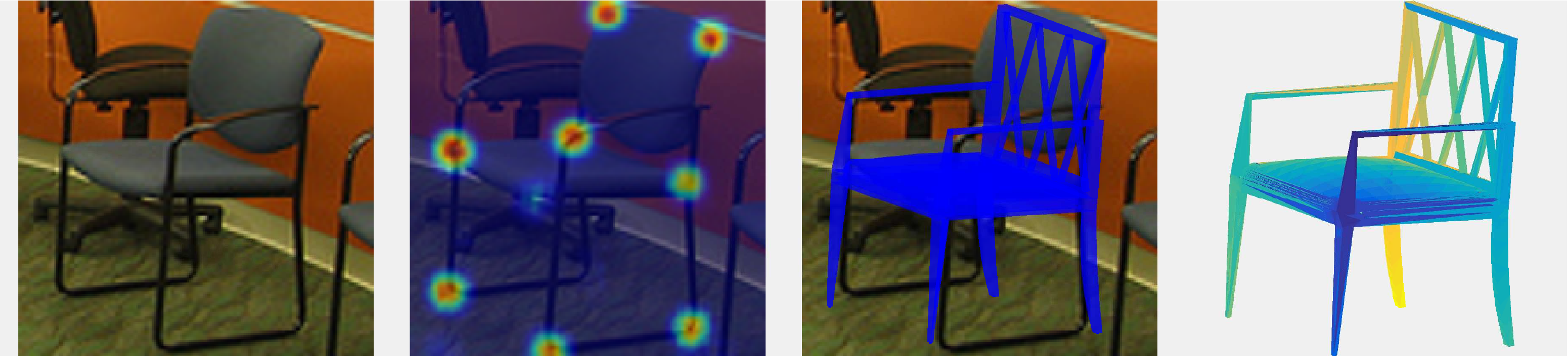}
  \includegraphics[width=0.49\linewidth]{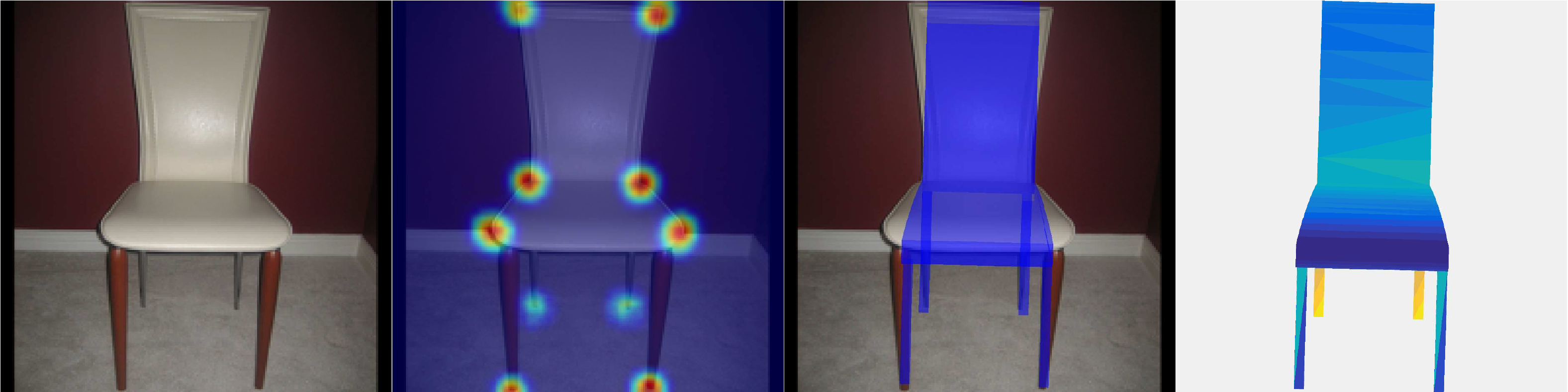} 
  \includegraphics[width=0.49\linewidth]{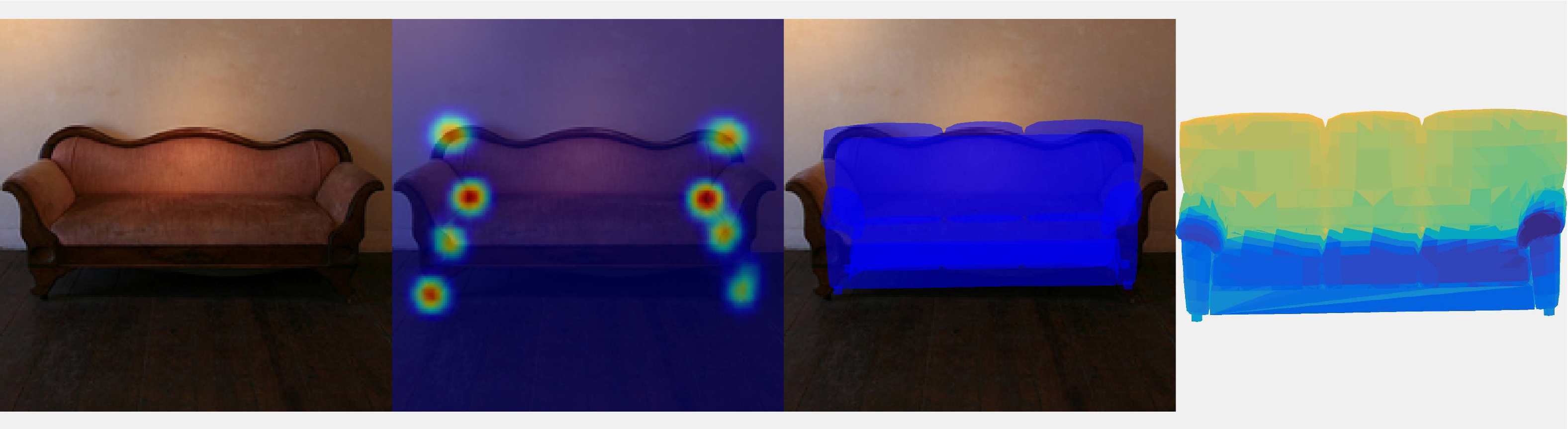} 
   \includegraphics[width=0.49\linewidth]{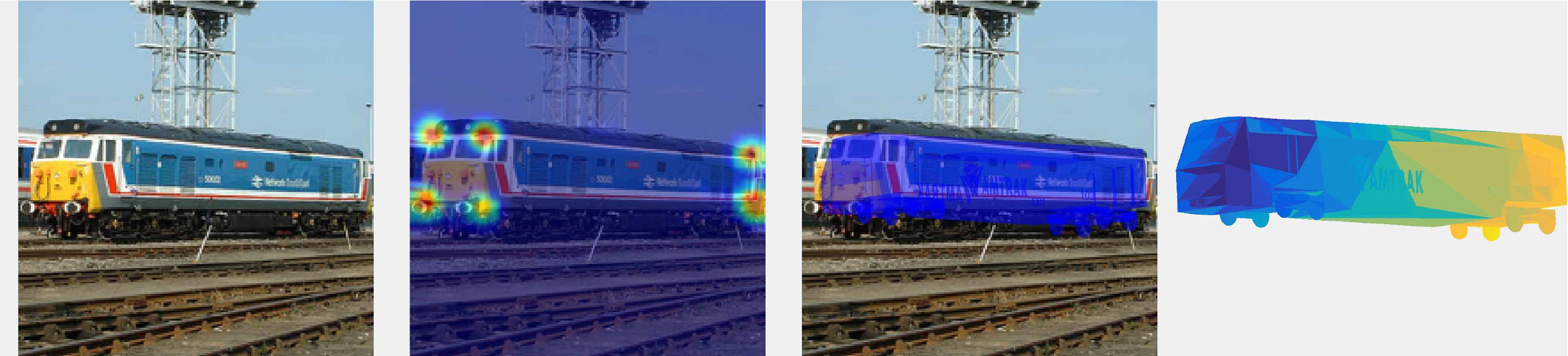}
   \includegraphics[width=0.49\linewidth]{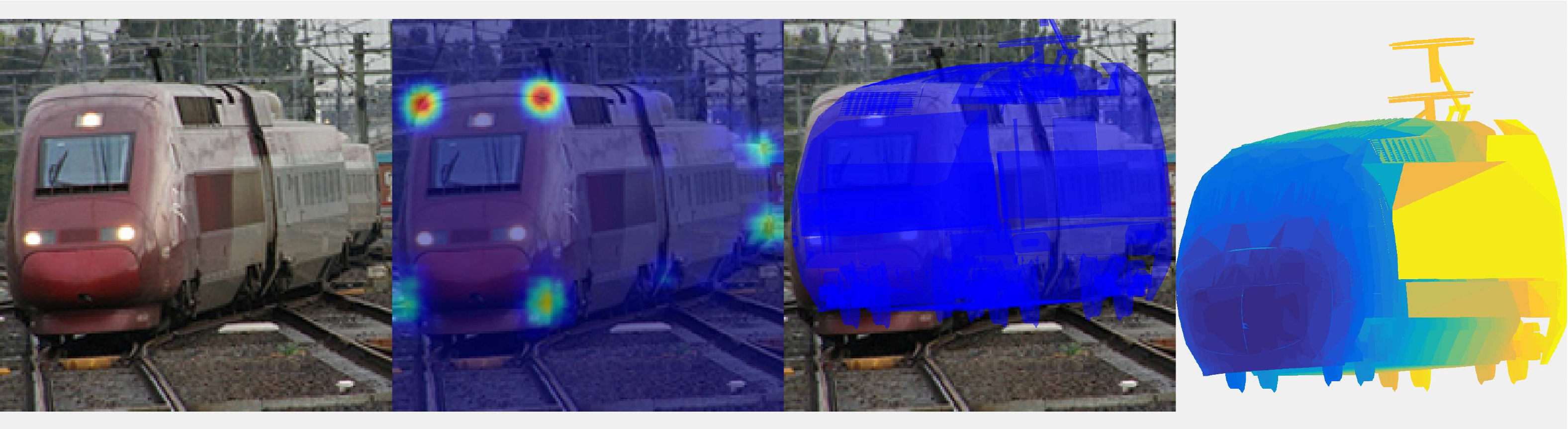}
  \caption{Example results of our approach on PASCAL3D+. For each example from left-to-right: 
the RGB image of the object, heatmap responses for the keypoints of the specific class, the CAD model projected to 2D after pose estimation, and the CAD model visualized in 3D.
}\label{fig:models3d}
\end{figure*}

\vspace{5pt}\noindent{\bf Pose estimation } The quantitative evaluation for pose estimation on PASCAL3D+ is presented in Table~\ref{tab:pascal3d}. Only the errors for rotations are reported as the 3D translation cannot be determined in the weak perspective case and the ground truth is not available as well. The rotational error is calculated using the geodesic distance, (\ref{eq:geodesic}). The proposed method shows improvement across most categories with respect to the state-of-the-art. The best results are achieved in the case where the fine subclass for the object is known and there exists an accurate CAD model correspondence. The proposed method with uniform weights for all keypoints is also compared as a baseline, which is apparently worse than considering the confidences during model fitting. A subset of results of the proposed method are visualized in Fig.~\ref{fig:models3d}.

\begin{table*}[t]
\caption{Viewpoint Estimation Median Error (degrees) on PASCAL3D+.}
\begin{center}
\begin{tabular}{c|cccccccccc}
\hline
   Approach    &     aero &       bike &  	 bottle &           bus &           car &         chair &	sofa &         train &         TV monitor  &	boat        \\
\hline
Tulsiani and Malik~\cite{tulsiani2015vk}	&         13.8 &         17.7 &     	12.9 &          5.8 &          9.1 &         14.8 &         15.2 &          8.7 &         {\bf 15.4} &    	{\bf 21.3}  \\
ours - PCA basis  &	11.2 &         15.2 &          13.1 &          4.7 &          6.9 &         12.7 &        21.7 &          9.1 &        38.5 &         37.9   \\
ours - CAD basis 	&   {\bf 8.0} &         {\bf 13.4} &        {\bf 11.7} &          {\bf 2.0} &          {\bf 5.5} &         {\bf 10.4} &          {\bf 9.6} &          {\bf 8.3} &          32.9 &         40.7 \\
ours - uniform weights 	&	16.3 &         17.8 &          14.1 &          11.7 &          30.7 &         17.6 &        32.4 &          20.8 &        25.0 &         72.0   \\
\hline
\end{tabular}
\end{center}
\label{tab:pascal3d}
\end{table*}

\begin{figure}
  \centering
   \includegraphics[width=\linewidth]{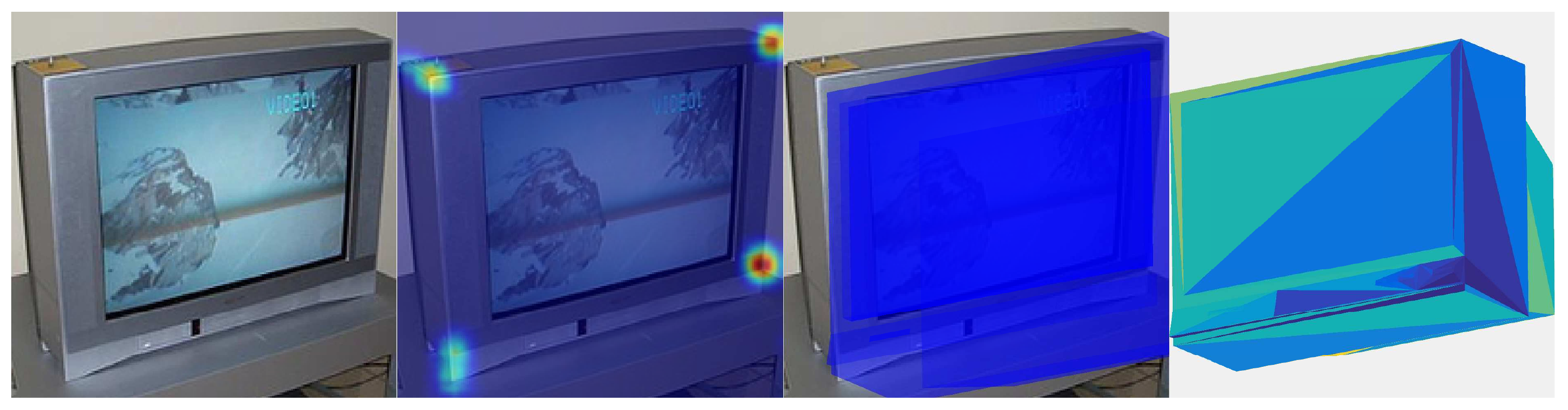}
   \includegraphics[width=\linewidth]{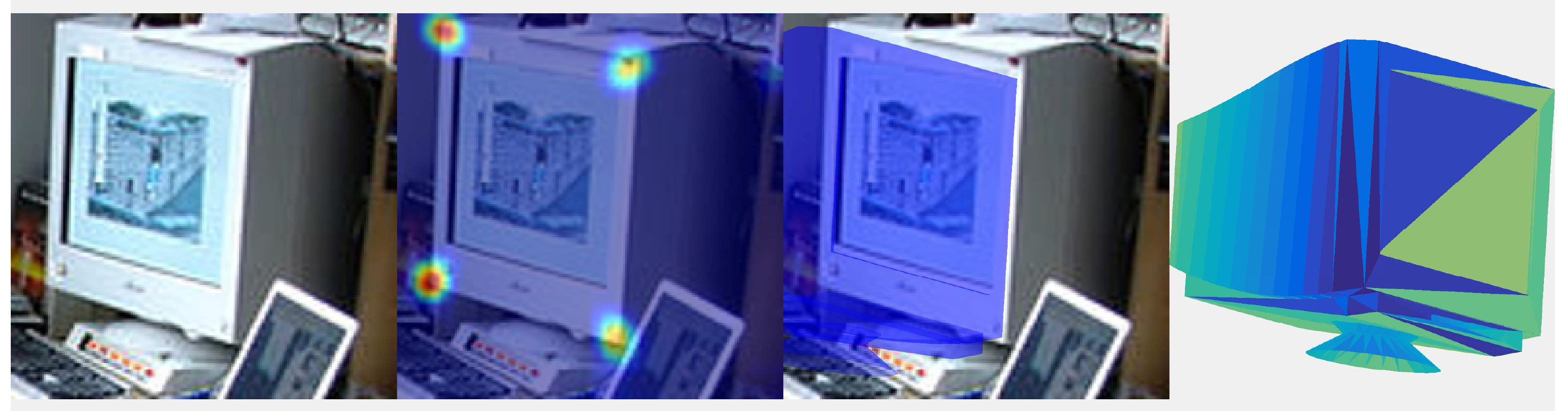}\\  
  \caption{Failure cases for the TV monitor class of PASCAL3D+. Although the visible keypoints are localized successfully the pose optimization fails because the 4 points are coplanar and the problem becomes ill-posed.}\label{fig:failure}
\vspace{-1em}
\end{figure}

\vspace{5pt}\noindent{\bf Failure cases } In Table~\ref{tab:pascal3d} we observed higher errors than the state-of-the-art for two classes, namely boat and TV monitor. For most images of TV monitor, there are only four coplanar keypoints. This makes pose estimation an ill-posed problem for the weak perspective case. Figure \ref{fig:failure} illustrates some failure cases because of this ambiguity. For boat we observed that in many cases the objects are very small and there are insufficient cues to discriminate between the front and the back, which makes the keypoint localization extremely hard. In these extreme cases, holistic and discrete viewpoint prediction might be more robust, which could in practice provide a prior to regularize our continuous prediction. We exclude the results for two classes from PASCAL3D+, namely dining table and motorbike, as we observed some inconsistency of the left-right definition in groundtruth annotations, which lead to erroneous training data. Since other approaches (e.g., \cite{tulsiani2015vk}) rely on discrete viewpoint annotations only, this issue is not reported.

\subsection{Processing time}

On a desktop with an Intel i7 3.4GHz CPU, 8G RAM and a GeForce GTX Titan X 6GB GPU, our pipeline needs around 0.2 seconds for the keypoint localization step and less than 0.1 seconds for the shape fitting step, for a total running time under 0.3 seconds. This makes our approach particularly suitable for applications where near real-time is desired. Moreover, further improvements in the running time are anticipated due to improvements in hardware, particularly with GPUs.

\section{Summary}
In this paper, we proposed an efficient method to estimate the continuous 6-DoF pose of an object from a single RGB image. Capitalizing on the robust semantic keypoint predictions provided by a state-of-the-art convnet, we proposed a pose optimization scheme that fits a deformable shape model to the 2D keypoints and recovers the 6-DoF pose of the object. To ameliorate the effect of false detections, our pose optimization integrates the heatmap response values in the optimization scheme to model the certainty of each detection. Both the weak perspective and the full perspective cases were investigated. The experimental validation included an instance-based scenario as well as full-scale evaluation on the PASCAL3D+ dataset, where we demonstrated state-of-the-art results for viewpoint estimation. Additionally, our method is accompanied by an efficient implementation with a running time under 0.3 seconds, making it a good fit for near real-time robotics applications.

\vspace{1em}

{\bf Acknowledgements:} We gratefully appreciate support through the following grants: NSF-DGE-0966142 (IGERT), NSF-IIP-1439681 (I/UCRC), NSF-IIS-1426840, ARL MAST-CTA W911NF-08-2-0004, ARL RCTA W911NF-10-2-0016, ONR N00014-17-1-2093, an ONR STTR (Robotics Research), NSERC Discovery, and the DARPA FLA program.


\bibliographystyle{IEEEtran}
\bibliography{bibref_definitions_short,IEEEexample}

\end{document}